\documentclass[letterpaper, 10 pt, conference]{IEEEtran}
\def\IEEEtitletopspace{16pt}

\usepackage{comment}
\usepackage{amsbsy}

\newcommand{\boldcheckmark}{\pmb{\checkmark}}

\IEEEoverridecommandlockouts                              

\usepackage[english]{babel}

\usepackage{graphicx}
\usepackage{animate}
\usepackage{epsfig} 				
\usepackage{epstopdf}

\usepackage{listings}
\usepackage{color}
\usepackage{nameref}

\usepackage{xcolor}
\usepackage[colorlinks,bookmarks=false,citecolor=blue,linkcolor=blue,urlcolor=blue]{hyperref}
\usepackage[numbers]{natbib}	 			

\usepackage{amsmath}	 			
\usepackage{amssymb}  				

\usepackage{dsfont}			
\usepackage{mathtools}

\usepackage{epigraph}
\usepackage{lscape}
\usepackage[]{nomencl}				
\usepackage{algorithm}
\usepackage{algorithmic}
\usepackage{multicol}
\usepackage{multirow}
\usepackage{makecell}
\usepackage{etoolbox}
\usepackage{graphics}
\usepackage{wrapfig}
\usepackage{xspace}

\usepackage{siunitx}
\usepackage{floatflt}
\usepackage{url}
\usepackage{placeins}
\usepackage{bm}

\usepackage{booktabs}
\usepackage{textcomp,mathcomp}
\usepackage{bm}

\usepackage{pifont}
\usepackage{gensymb}

\newcommand{\link}[1]{\colora{\url{#1}}}

\newcommand{\fig}[1]{Fig.~\ref{#1}}
\newcommand{\eq}[1]{Equation~\eqref{#1}}

\newcommand{\tab}[1]{Table~\ref{#1}}

\newcommand{\ProjectWeb}[0]{\href{https://linchangyi1.github.io/LocoMan}{https://linchangyi1.github.io/LocoMan}}

\newcommand{\robot}[0]{LocoMan\xspace}
\newcommand{\manipulator}[0]{loco-manipulator\xspace}
\newcommand{\manipulators}[0]{loco-manipulators\xspace}

\renewcommand{\vec}{\bm}

\title{\LARGE \bf \robot: Advancing Versatile Quadrupedal Dexterity with \\Lightweight Loco-Manipulators}

\author{Changyi Lin$^{1}$, Xingyu Liu$^{1}$, Yuxiang Yang$^{2}$, Yaru Niu$^{1}$, \\Wenhao Yu$^{3}$, Tingnan Zhang$^{3}$, Jie Tan$^{3}$, Byron Boots$^{2}$, Ding Zhao$^{1}$\\
\ProjectWeb
\thanks{$^{1}$ Carnegie Mellon University}
\thanks{$^{2}$ University of Washington}
\thanks{$^{3}$ Google Deepmind}
}

\usepackage[font=footnotesize]{caption}

\begin{document}

\maketitle


\begin{abstract}
	Quadrupedal robots have emerged as versatile agents capable of locomoting and manipulating in complex environments. Traditional designs typically rely on the robot's inherent body parts or incorporate top-mounted arms for manipulation tasks. However, these configurations may limit the robot's operational dexterity, efficiency, and adaptability, particularly in cluttered or constrained spaces. In this work, we present \robot, a dexterous quadrupedal robot with a novel morphology to perform versatile manipulation in diverse constrained environments.
By equipping a Unitree Go1 robot with two low-cost and lightweight modular 3-DoF \manipulators on its front calves, \robot leverages the combined mobility and functionality of the legs and grippers for complex manipulation tasks that require precise 6D positioning of the end effector in a wide workspace. To harness the loco-manipulation capabilities of \robot, we introduce a unified control framework that extends the whole-body controller (WBC) to integrate the dynamics of \manipulators. Through experiments, we validate that the proposed whole-body controller can accurately and stably follow desired 6D trajectories of the end effector and torso, which, when combined with the large workspace from our design, facilitates a diverse set of challenging dexterous loco-manipulation tasks in confined spaces, such as opening doors, plugging into sockets,
picking objects in narrow and low-lying spaces, and bimanual manipulation.

\end{abstract}


\section{Introduction}
\label{sec:intro}

Recent advances in the capability of quadrupedal robots have enabled them to traverse highly complex terrains \cite{jenelten2024dtc, choi2023learning, lee2020learning, yang2023neural, kumar2021rma, lindqvist2022multimodality}, perform acrobatic skills \cite{yang2023cajun, li2023learning,zhuang2023robot,cheng2023extreme, hoeller2023anymal}, and interact with humans \cite{xiao2021robotic}.
While most of these works focus on improving the \emph{mobility} of quadrupedal robots, the \emph{manipulation} capability of quadrupedal robots still remains limited and often requires special designs to achieve tasks such as pressing buttons\cite{cheng2023legs} and kicking balls\cite{ji2022hierarchical, ji2023dribblebot}.
To be effectively deployed in daily life, quadrupedal robots are required to possess versatile manipulation skills with enhanced dexterity, in addition to their inherent locomotion capability.

Integrating manipulation into quadrupedal robots proves to be a difficult task. As illustrated in Table.~\ref{tab:morphology_comparison}, 
without external hardware modifications, quadrupedal robots have to utilize their legs, head, or torso to move or transport objects \cite{cheng2023legs, ji2023dribblebot, jeon2023learning, sombolestan2023hierarchical}, which limits their ability to control the 6D poses of the object and perform high-precision tasks.
A popular solution to address this challenge is to use a top-mounted robot arm \cite{fu2023deep}.
However, this solution often comes at the cost of increased payload and decreased agility, and prevents the robot from reaching narrow spaces or performing more dexterous tasks such as bimanual manipulation.
While foot-mounted grippers offer a lightweight alternative to full robot arms, existing work adopts 1-DoF grippers with limited functionality~\cite{tsvetkov2022novel}.
Therefore, it can be challenging to achieve versatile dexterous loco-manipulation using existing manipulator solutions for legged robots.

\begin{figure}[t]
	\centering
	\includegraphics[width= \linewidth]{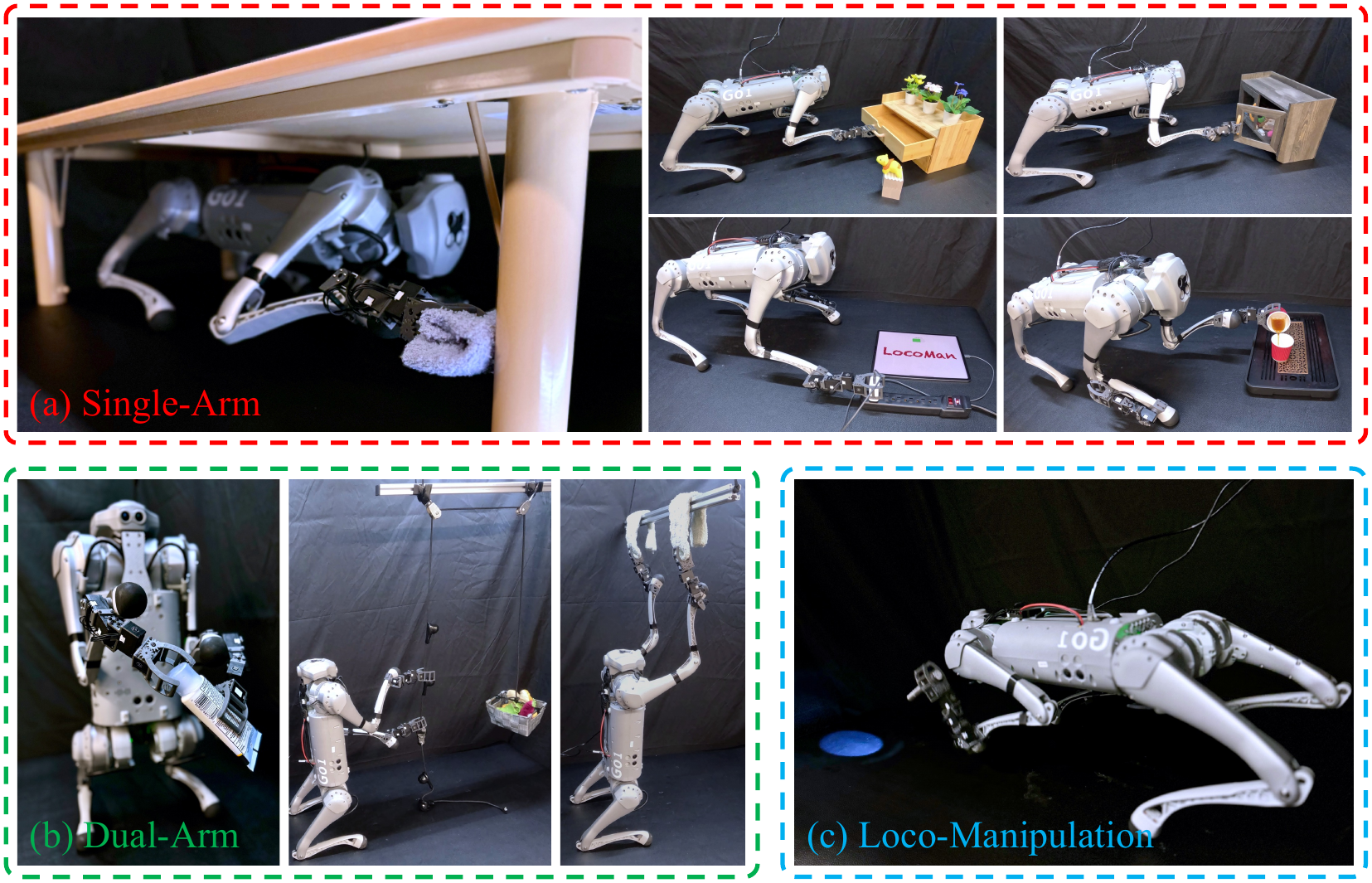}
	\caption{Equipped with \manipulators, \robot is proficient in handling versatile manipulation tasks. (a) With a single \manipulator, \robot not only can perform manipulation tasks that require precision and stability, but also excels at operating in extremely narrow space in its compact form. (b) With both \manipulators installed on the two front legs, \robot is able to perform bimanual manipulation tasks when standing upright. (c) \robot is also capable of loco-manipulation, e.g. carrying objects with its \manipulators while walking.}
	\label{fig:teaser}
    \vspace{-9pt}
\end{figure}

In this work, we present \robot, a comprehensive solution for versatile and skilled manipulation with legged robots.
To reduce the payload and increase dexterity, \robot mounts two custom-designed lightweight 3-DoF manipulators on the front \emph{calves} of the robot, and uses existing leg joints in addition to the manipulators for manipulation, as illustrated in \fig{fig:teaser}. We refer to our designed manipulator as ``loco-manipulator'', which draws conceptual inspiration from the front limbs of animals such as great apes and bears, which are adept at both locomotion and skilled manipulation.

For precise and robust control of \robot to perform loco-manipulation tasks, we design a unified whole-body control framework, which tracks the command kinematically and dynamically through joint impedance commands.
The integration of the \manipulator's lightweight design with precise whole-body control enables \robot to operate in diverse modes, from reaching an object from narrow-space with single-arm, bimanual manipulation while sitting upright, to carrying an object while walking.
We further design a Finite State Machine (FSM) to manage the transitions of these operation modes.

We evaluate the design of \manipulator on a Unitree Go1 robot and find that the added \manipulators can significantly increase the robot's workspace (by 99.01\% for single-arm manipulation and 118.28\% for bimanual manipulation) with negligible weight increase ($<$2.5\%).
In addition, the whole-body controller enables precise trajectory tracking with a mean pose error of \text{$1.89$} $mm$ and \text{$0.047$} $rad$.
We further validate the capabilities of \robot in a number of real-life manipulation tasks under user teleoperation.
In addition to standard tasks such as cabinet opening and drink pouring, \robot can reach narrow spaces with limited vertical clearance (\text{$0.25$} $m$ for locomotion and \text{$0.09$} $m$ for manipulation).
Thanks to the lightweight design of \manipulator, \robot can stand on its rear legs in an upright pose and perform bimanual manipulation tasks such as grasping a pair of socks simultaneously and hoisting a basket with a rope.

In summary, the contribution of this paper is as follows:
\begin{enumerate}
    \item We present \robot, leg-mounted with two lightweight \manipulators to improve manipulation capability of quadrupedal robots.
    \item We design a unified whole-body controller for precise, simultaneous 6D pose tracking for both the end effectors and the torso.
    \item We evaluate that \robot is capable of precise trajectory tracking in a large workspace.
    \item We demonstrate that \robot can perform versatile real-life manipulation tasks, including cabinet opening, charger inserting, drink pouring, grasping in a narrow space, and basket hoisting.
\end{enumerate}

\begin{table}[t]
\setlength{\tabcolsep}{1pt}
\centering
\caption{Comparison of the capabilities between robots with different morphology.}
\begin{tabular}{c|c|c|c|c}\toprule
Methods & \cite{cheng2023legs,ji2023dribblebot,arm2024pedipulate} & \cite{fu2023deep,sleiman2021unified} & \cite{tsvetkov2022novel} & \textbf{Ours} \\
\midrule
\multirow{2}{*}{Additional Manipulator} & \multirow{2}{*}{None} & \multirow{2}{*}{\begin{tabular}[c]{@{}c@{}} 6DoF Top- \\ mounted \end{tabular}} & \multirow{2}{*}{\begin{tabular}[c]{@{}c@{}} 1DoF Leg- \\ mounted \end{tabular}} & \multirow{2}{*}{\begin{tabular}[c]{@{}c@{}} 3DoF Leg- \\ mounted \end{tabular}}  \\ 
&&&& \\
\midrule
6D Operational Space & \ding{55} & \boldcheckmark & \ding{55} & \boldcheckmark \\
Narrow Space Manipulation & \ding{55} & \ding{55} & \boldcheckmark & \boldcheckmark \\
Bi-Manual Manipulation & \ding{55} & \ding{55} & \boldcheckmark & \boldcheckmark \\
Loco-Manipulation & \boldcheckmark & \boldcheckmark & \ding{55} & \boldcheckmark \\
\bottomrule
\end{tabular}
\label{tab:morphology_comparison}
\end{table}

 

\section{Related Work}
\label{sec:related}

	\subsection{Quadrupedal Loco-Manipulation}

\textbf{Without Additional Manipulators} Researchers have a long history exploring the manipulation capability of quadrupedal robots.
Without an additional manipulator, researchers have designed the robot to use its head \cite{sombolestan2023hierarchical, sombolestan2023hierarchical1, nachum2019multi, mataric1995cooperative}, torso \cite{jeon2023learning}, or foot \cite{cheng2023legs, ji2022hierarchical, ji2023dribblebot, shi2021circus, kolvenbach2019haptic, topping2017quasi, Kim-RSS-22} to interact with objects in the environment, where \cite{sombolestan2023hierarchical1, nachum2019multi, mataric1995cooperative} rely on multiple robots to push larger objects.
However, without an additional mechanism such as a gripper, these approaches are limited to relatively simple tasks that do not require fetching objects, such as kicking balls \cite{ji2022hierarchical}, pressing buttons \cite{cheng2023legs}, pushing boxes \cite{mataric1995cooperative}, opening doors \cite{topping2017quasi}, and probing \cite{kolvenbach2019haptic}.

\textbf{With Additional Manipulators} Many previous works mount a dedicated arm on top of the quadrupedal robot to perform more complicated manipulation tasks such as pick-and-place \cite{yokoyama2023adaptive, chiu2022collision}, pulling doors \cite{sleiman2023versatile, sleiman2021unified, mittal2022articulated}, tablecloth spreading \cite{chu2023model}, turning wheels \cite{ferrolho2023roloma}, wiping a whiteboard \cite{fu2023deep}, and collaboratively carrying large objects \cite{de2023centralized, kim2022cooperative}. On the other hand, equipping a robot with a dedicated arm typically adds to the mechanical complexity and power requirements, leading to an increase in both weight and expense. Moreover, such an addition makes it difficult to operate in a narrow space and reduces the quadrupedal robot's agility.
Some other works alleviate these problems by installing more lightweight manipulators or grippers on the mouth \cite{lykov2024cognitivedog} or foot \cite{tsvetkov2022novel, arm2024pedipulate} of the quadrupedal robot.
Tsvetkov et al. \cite{tsvetkov2022novel} propose a novel design of a small-scale quadrupedal robot with manipulators built into the legs, each requiring three additional actuators, where two are used for grasping. It facilitates single-arm and two-arm manipulation, but it only supports specifying target positions in joint space.
Arm et al. \cite{arm2024pedipulate} propose a learning-based controller that can track quadrupedal foot position targets through whole-body motions, and a 2-DoF gripper is attached to the foot to perform tasks such as collecting rock samples. However, how to efficiently reach a specified 6D pose of the end effector on the leg to perform more complex manipulation tasks is still underexplored.
While some hexapod robots \cite{roennau2014lauron, heppner2014versatile, heppner2015laurope, whitman2017generating, brinkmann2020enhancement} use legs as manipulators, where integrated grippers can fetch objects with high degrees of freedom, prehensile manipulation with dexterity could be challenging for quadrupedal robots.

\subsection{Controllers for Leg-Manipulation}

To accomplish a wide range of manipulation tasks, the controller of the loco-manipulation robot need to support a variety of operation modes, including reaching a narrow space with a single arm, walking while carrying an object, and bi-manual manipulation with upright pose.
While prior works have achieved similar tasks such as grasping \cite{sleiman2021unified, sleiman2023versatile, fu2023deep}, object pushing and kicking \cite{sombolestan2023hierarchical, ji2022hierarchical, ji2023dribblebot}, and object carrying \cite{whitman2017generating} using optimization-based \cite{sleiman2021unified, sleiman2023versatile, yokoyama2023adaptive, sombolestan2023hierarchical, whitman2017generating} or learning-based \cite{jeon2023learning, ji2022hierarchical, ji2023dribblebot, fu2023deep, cheng2023legs} methods, most of these controllers are designed for specific tasks, and additional effort is required for multi-task support.
In contrast, \robot adopts a unified control framework, which uses the same low-level whole-body controller \cite{kim2019highly} for all five operation modes.
By tracking desired trajectories at both the kinematics and dynamics level, our unified framework requires little task-specific tuning and performs all manipulation tasks with high accuracy.


\section{Design of \manipulator} 
\label{sec:design}

	\begin{figure}[t]
	\centering
	\includegraphics[width= \linewidth]{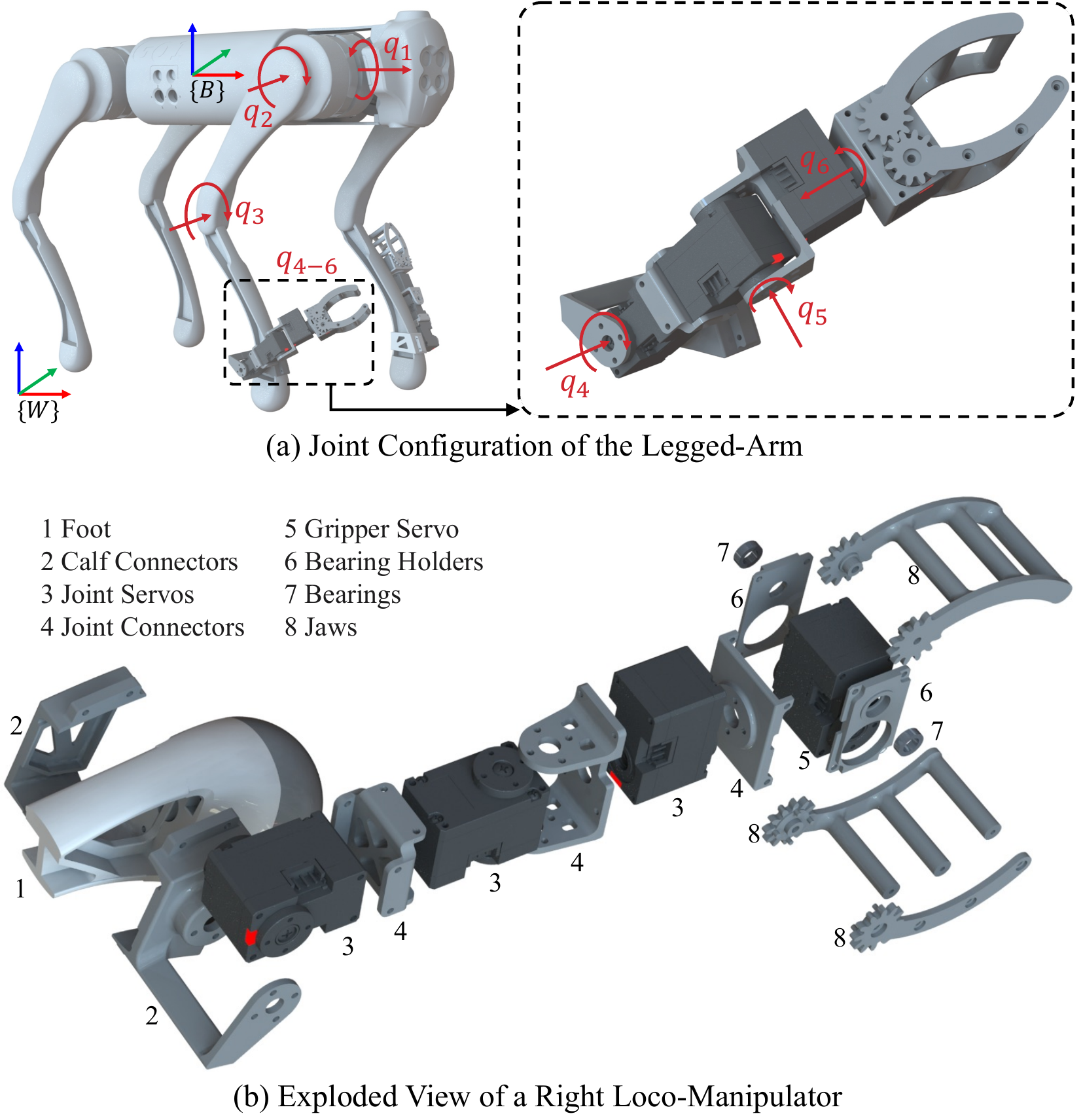}
	\caption{Design of our \manipulator. (a) The joint configuration of the legged-arm. The legged-arm has six joints including three from the leg and three from \manipulator. (b) The components of a right \manipulator shown in an exploded view.}
	\label{fig:design}
    \vspace{-10pt}
\end{figure}

\begin{table*}[ht]
\centering
\caption{Specifications of a \manipulator.}
\begin{tabular}{ccccccc}\toprule
 DoFs & Dimension [$mm^3$] & Weight [$g$] & Cost[\$] & 4 Servos & 2 Bearings [$d\times D \times h$] & 3D Printer\\
\midrule
3 (joints) + 1 (gripper)  & $179\times61\times42$& 147 & 370 & XC330-T288-T &\text{$5$} $mm\times\text{8}$ $mm\times\text{2.5}$ $mm$ & Formlabs\textsuperscript{\textregistered} Form 3+\\
\bottomrule
\end{tabular}
\label{tab:design}
\end{table*}

To accomplish versatile manipulation tasks, the robot must be able to reach desired 6D poses in its workspace.
While quadrupedal robots can utilize torso movements to aid manipulation \cite{arm2024pedipulate, tsvetkov2022novel}, enabling 6D pose reaching directly from the manipulator can significantly improve the precision and range of manipulation tasks.
Combining the loco-manipulator DoFs with the existing joint DoFs, each front limb of \robot has 6 DoFs in total, and can reach arbitrary 6D space poses \emph{without} body movement.

The design of \robot is inspired by the anatomy of human arms.
As illustrated in~\fig{fig:design}(a), we attach custom-designed \manipulators to the calf of the front legs of a quadrupedal robot, and utilize existing leg joints ($q_{1-3}$) together with the gripper joints ($q_{4-6}$) for 6DoF space manipulation.
The original leg joints, $q_{1-3}$, similar to human shoulders and elbows, are used mainly for movement and positional tracking.
The added joints of \manipulator, ($q_{4-6}$), similar to the human wrist, are primarily used for orientation tracking.
Integrated together, each front limb of \robot can reach a wide variety of space poses effectively.

We carefully orient the joints of \manipulator to seamlessly integrate with the existing structure of the original robot. 
Specifically, we design the first joint ($q_4$) to be aligned with the calf joint of the front leg, design the last joint ($q_6$) to point towards the gripper, and design the second joint ($q_5$) to be perpendicular to the other two joints.
In this way, the gripper can reach out to far spaces during manipulation (front-right foot of \fig{fig:design}(a)), while staying tucked to the calf during standing and walking (front-left foot of \fig{fig:design}(a)) with minimal interference with locomotion capabilities.

The components of \manipulator are lightweight, low cost, and easy to fabricate. As illustrated in \fig{fig:design}(b),
for the actuators of a \manipulator, we select four Dynamixel servos, which are compact, highly dynamic, and capable of providing position, velocity, and torque feedback.
\manipulator is mounted on the robot's calf via two calf connectors, which are meticulously designed so that their internal structures seamlessly interlock with the skeletal framework of the robot's calf.
To make \manipulator more compact, we designed the gripper as two pairs of rotating jaws with gear engagement, enabling symmetric opening and closing of the gripper with a single servo. One pair of jaws is connected to the servo horns, while the other pair is mounted with two bearings.
Except for the servos (label 3 and label 5 in~\fig{fig:design}(b)) and the bearings (label 7), the other components of \manipulator, including the calf connectors, joint connectors, bearing holders, and jaws, can be 3D printed using a standard 3D printer.
Integrated together, each \manipulator weighs \text{$147$}$g$ ($1.23\%$ of the robot's weight), and can be easily fabricated with a material cost of $\$370$ (\tab{tab:design}).


\section{A Unified Framework for Whole-Body Loco-Manipulation}
\label{sec:whole_body}

	\begin{figure*}[ht]
    \centering
    \includegraphics[width=\linewidth]{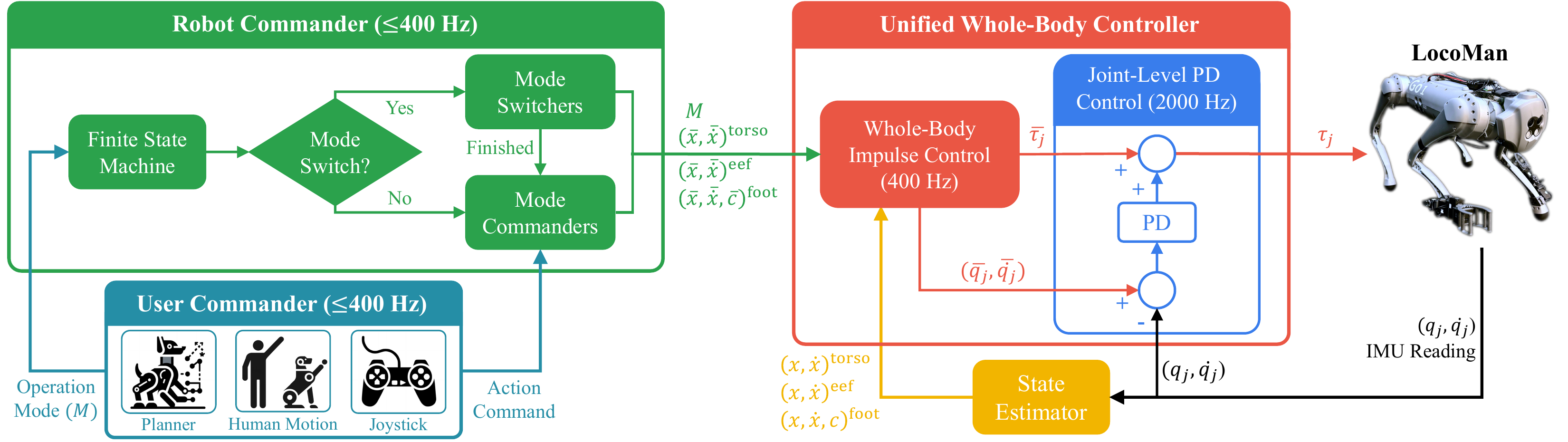}
    \caption{The unified framework for whole-body loco-manipulation. The robot commander converts the desired action command from user into the standardized robot command based on a specific operation mode ($M$). The unified whole-body controller, including whole-body impulse control and joint-level PD control, computes torques for each joint of \robot to track the desired robot command,}
    \label{fig:pipeline}
    \vspace{-3pt}
\end{figure*}

\begin{table*}[ht]
\centering
\caption{Hierarchical tracking objectives and state estimators for the five operation modes. The subscripts $\{m\}$, $\{s\}$, $\{l, r\}$ indicate the end effector in manipulation ($m$), the feet in swing ($s$), and the left ($l$) and right ($r$) grippers respectively.}
\begin{tabular}{cc|c|c|cc}\toprule
\multicolumn{1}{c}{\multirow{2}{*}{Operation Mode}} & \multicolumn{4}{c}{Hierarchical Tracking Objectives in WBC} & \multicolumn{1}{c}{\multirow{2}{*}{State Estimator}}\\
\cmidrule(lr){2-5}
~ & Objective 1 & Objective 2 & Objective 3 & Objective 4 & ~\\
\midrule
Single Foot Manipulation & Torso Position & Torso Orientation & Foot$_m$ Position & - & Kinematics-Based\\
Single Gripper Manipulation & Torso Position & Torso Orientation & Gripper$_m$ Position & Gripper$_m$ Orientation & Kinematics-Based\\
Bimanual Manipulation & Gripper$_{l,r}$ Positions & Gripper$_{l,r}$ Orientations & - & - & Kinematics-Based\\
Locomotion & Torso Velocity &  Torso Orientation & Foot$_s$ Positions & - & Kalman Filter-Based\\
Loco-Manipulation& Torso Velocity &  Torso Orientation & Foot$_s$ Positions & Gripper$_{l,r}$ Orientations & Kalman Filter-Based\\
\bottomrule
\end{tabular}
\label{tab:operation_modes}
\vspace{-7pt}
\end{table*}

\subsection{Notation}
\label{notation}
We denote the state of \robot as $\vec{q}=(\vec{x}^{\text{torso}}, \vec{q}_{j})\in \mathbb{R}^{24}$, where $\vec{x}^{\text{torso}}\in\mathbb{R}^{6}$ denotes the position and orientation of the floating torso base, and $\vec{q}_{j} \in\mathbb{R}^{18}$ denotes the joint angles of \robot, including the legs' ($\text{12}$ DoF) and the \manipulators' ($\text{6}$ DoF). We denote $\vec{x}^{\text{eef}}=(\vec{x}^{\text{eef}}_{\text{left}}, \vec{x}^{\text{eef}}_\text{right})\in \mathbb{R}^{12}$ and $\vec{x}^{\text{foot}}\in \mathbb{R}^{24}$ as the cartesian positions and orientations of two grippers and four feet, which can be computed from the state vector using forward kinematics. 
We use the bar notation $(\overline{\cdot})$ to denote all desired states.

\subsection{Overview}
Effective loco-manipulation requires the robot to complete a large variety of tasks accurately and robustly, from in-place picking to locomotion while carrying objects.
In light of this requirement, we design a unified control framework for \robot (Fig.~\ref{fig:pipeline}), where a \emph{robot commander} specifies the desired state of \robot based on the action command from user, and a \emph{unified whole-body controller} (WBC) tracks the desired state based on \robot's kinematics and dynamics model.
The robot commander supports a variety of operation modes (\tab{tab:operation_modes}), such as locomotion and in-place manipulation, where the mode switch is governed by a finite state machine (FSM).
Within each mode, the FSM launches a mode-specific \emph{mode commander} to convert user specified target (e.g. walking speed or gripper pose) into desired states of the torso $(\overline{\vec{x}}, \overline{\dot{\vec{x}}})^{\text{torso}}$, end effector $(\overline{\vec{x}}, \overline{\dot{\vec{x}}})^{\text{eef}}$, and foot $(\overline{\vec{x}}, \overline{\dot{\vec{x}}}, \overline{\vec{c}})^{\text{foot}}$.
We further design a set of \emph{mode switchers} to smoothly transition between different operation modes.
For robust operation and accurate tracking of the desired state, we extend a \emph{whole-body controller} to solve the desired position $\overline{\vec{q}_{j}}$, velocity $\overline{\dot{\vec{q}}_{j}}$ and torque $\overline{\vec{\tau}_{j}}$ of each joint based on the complete kinematics and dynamics model of the robot. Then a joint-level PD controller computes the final torque $\vec{\tau_{j}}$ for each joint at a higher frequency.
In addition, we implement two state estimators to accurately estimate the robot state based on sensor feedback.

\subsection{Robot Commander}
The \emph{robot commander} supports five different operation modes (\tab{tab:operation_modes}), namely locomotion, locomanipulation, single-foot manipulation, single-gripper manipulation, and bimanual manipulation. 
Each \emph{mode commander} converts different user commands from planners, human motions, or joysticks to desired states of \robot.

For single-arm manipulation, the user specifies the desired torso 6D pose, and the desired gripper 6D pose or foot 3D position. For bimanual manipulation, the user commands the desired 6D poses for each gripper. The stance feet remain fixed to the ground during the three manipulation modes.
For locomotion and loco-manipulation, the user specifies the desired velocity, roll, pitch, and height of the robot torso, together with the desired gripper orientation for the later mode.

To ensure a smooth transition between operation modes, the FSM launches a specific \emph{mode switcher} during mode transitions.
For example, from locomotion mode to single-arm manipulation mode, \robot is commanded to zero velocity until all feet are on the ground, and then moves the torso to the rear opposite the end effector being manipulated.

\subsection{Unified Whole-Body Controller}
Given the desired states of the torso $(\overline{\vec{x}}, \overline{\dot{\vec{x}}})^{\text{torso}}$, end effector $(\overline{\vec{x}}, \overline{\dot{\vec{x}}})^{\text{eef}}$, and foot $(\overline{\vec{x}}, \overline{\dot{\vec{x}}})^{\text{foot}}$ and foot contact force $ \overline{\vec{c}}^{\text{foot}}$, the WBC computes the desired position $\overline{\vec{q}_{j}}$, velocity $\overline{\dot{\vec{q}}_{j}}$ and torque $\overline{\vec{\tau}_{j}}$ of each joint.

Our WBC implementation is based on the two-step implementation from \citet{kim2019highly} with additional modeling of the loco-manipulators.
In the first step, WBC tracks the desired states \emph{kinematically} by computing the desired position $\overline{\vec{q}_{j}}$ and velocity $\overline{\dot{\vec{q}}_{j}}$ of each joint using Inverse Kinematics (IK).
Since the robot may not have sufficient degrees of freedom to track the entire desired state, we assign a \emph{objective priority} to each component of the desired states (\tab{tab:operation_modes}), and solve objectives in descending priority by projecting low-priority objectives into the null-space of high-priority ones.
In the second step, WBC tracks the desired state \emph{dynamically} by optimizing joint torque commands $\overline{\vec{\tau}_{j}}$, where the objective is to track the desired torso acceleration, and the constraints include rigid body dynamics and friction cone limits.
Please refer to the original work \cite{kim2019highly} for additional details.


\section{Experiments}
\label{sec:experiments}

	We design a set of experiments to validate the capability of \robot in versatile manipulation tasks. Specifically, we aim to answer the following questions:
(1) How large is the workspace of \robot and how does it compare with the workspace of the original robot?
(2) How well can \robot track a desired trajectory in its workspace?
(3) What real-life manipulation tasks can \robot perform? and
(4) What is the benefit of the leg-mounted manipulator design of \robot, compared to the top-mounted arms~\cite{fu2023deep,sleiman2021unified}?

\subsection{Experiment Setup}
As illustrated in~\fig{fig:teaser}, we install two loco-manipulators on the front legs of Unitree Go1, a quadrupedal robot. The servos of the manipulators are powered by the robot via an additional voltage converter (from 24V to 12V). The algorithms of the motion command and control framework run on a desktop, and the control signals for the quadrupedal robot's motors and the \manipulators' servos are sent via cables.

\subsection{Workspace Analysis}

\begin{figure}[t]
	\centering
	\includegraphics[width= 0.98\linewidth]{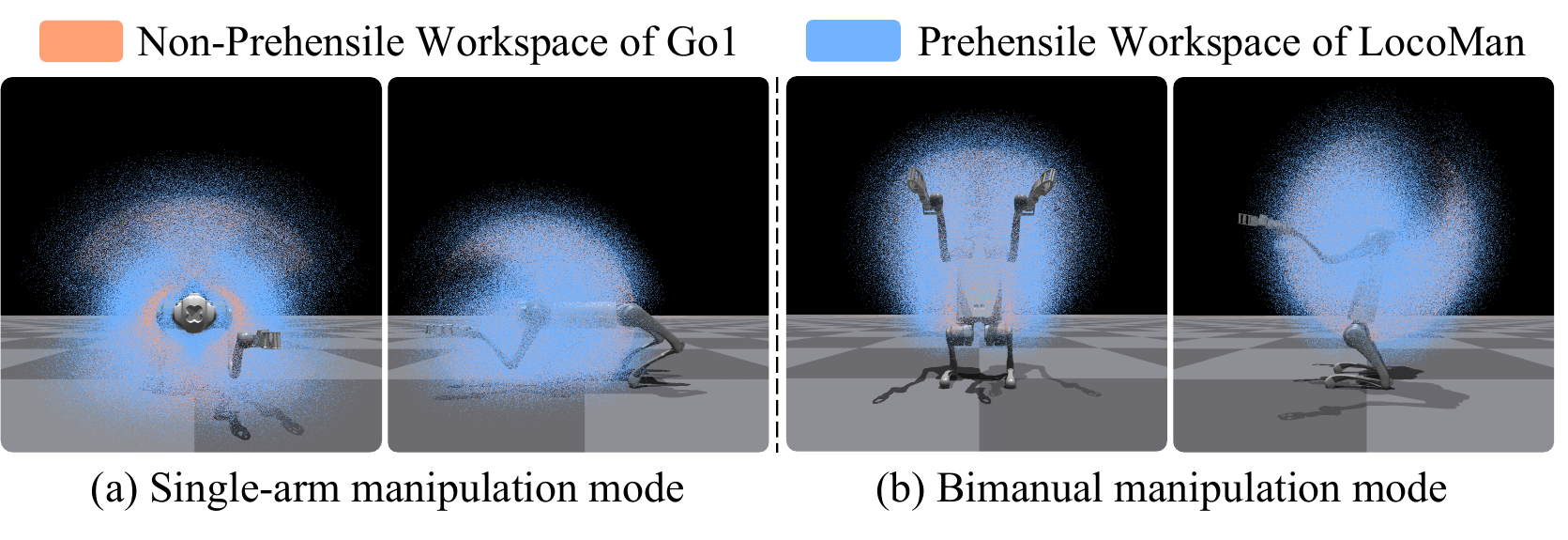}
	\caption{Our loco-manipulator turns the original non-prehensile (in orange) workspace of Go1 to a prehensile workspace (in blue) and expand the reachable area by more than 80\%.}
	\label{fig:workspace}
\end{figure}

We compare the workspace of \robot with the workspace of the unmodified Unitree Go1 robot as illustrated in \fig{fig:workspace}.
The workspace is approximated by uniformly sampling multiple angles for each joint within its joint limit, recording the joint configurations without self-collision, and plotting the end effector positions (center of the gripper for \robot and foot for the original robot).
With additional \manipulators, \robot extends the workspace of the end effector much further from the original foot toes to the prehensile gripper.
Quantitatively, the inclusion of the loco-manipulator expands the workspace volume by at least 80\%: from \text{$0.34$}$m^3$ to \text{$0.68$}$m^3$ in the single-arm manipulation mode, from \text{$0.38$}$m^3$ to \text{$0.83$}$m^3$ in the bimanual mode, and from \text{$0.63$}$m^3$ to \text{$1.14$}$m^3$ in the combined volume.

\subsection{Trajectory Tracking}
\begin{figure}[t]
	\centering
	\includegraphics[width= \linewidth]{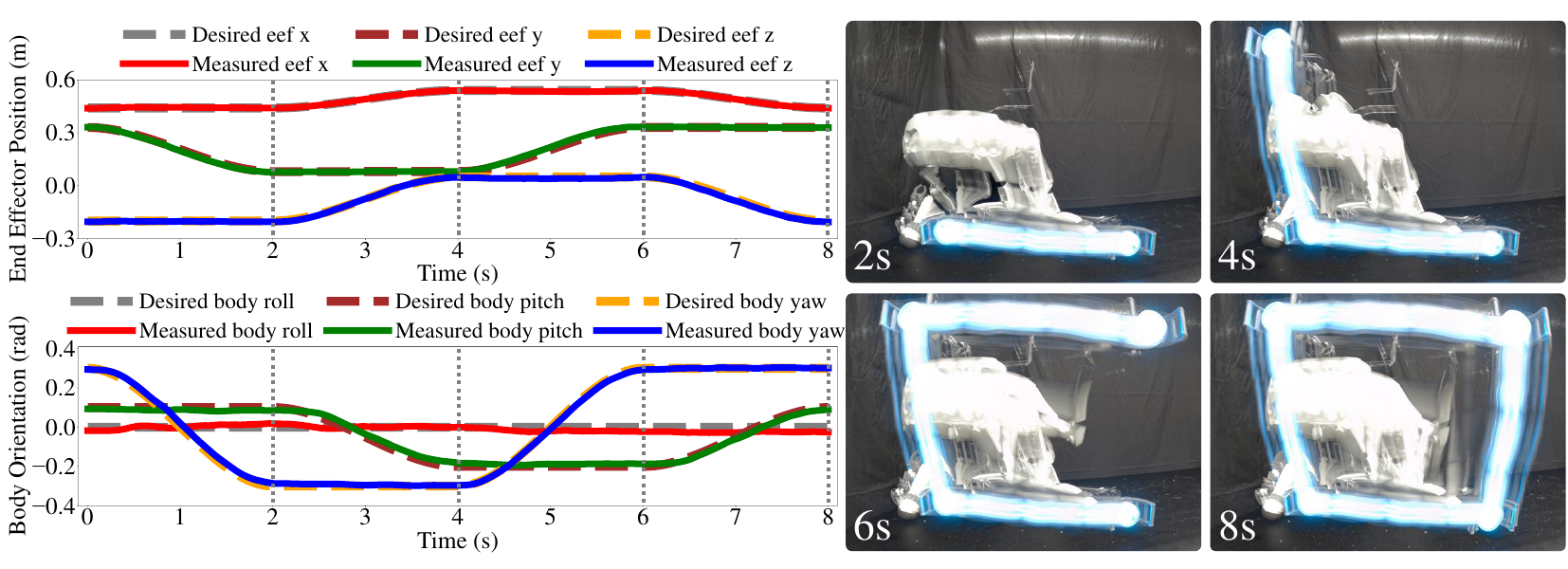}
	\caption{The figure on the left shows trajectory tracking of the end effector position and the torso orientation when drawing a spatial square. The figure on the right depicts the corresponding 3D trajectory tracked by an LED light attached to the gripper.}
	\label{fig:eef_tracking}
    \vspace{-9pt}
\end{figure}

\begin{table}[t]
\centering
\caption{Translational and rotational MAEs of end effector and torso in trajectory tracking.
Unit is $mm$ for lengths and $rad$ for angles.}
\begin{tabular}{cccccc}\toprule
Trajectories &  ${e}_{\text{eef}}^t$ &  ${e}_{\text{eef}}^R$ &  ${e}_{\text{torso}}^t$ &  ${e}_{\text{torso}}^R$ \\
\midrule
Gripper-M w/o Torso Motion & 1.67 & 0.045 & 1.92 & 0.0075 \\
Gripper-M w Torso Motion & 1.89 & 0.047 & 2.38 & 0.016 \\
Foot-M w/o Torso Motion & 0.93 & - & 2.02 & 0.0078 \\
Foot-M w Torso Motion & 1.10 & - & 2.26 & 0.017 \\
Standing w Torso Motion & - & - & 1.05 & 0.014 \\
\bottomrule
\end{tabular}
\label{tab:tracking_error}
\vspace{-6pt}
\end{table}

We design a set of experiments to evaluate the trajectory tracking performance of \robot.
First, we manually design trajectories of the gripper and torso, where the gripper follows a desired  rectangular trace with a spatial bounding box of \text{0.25}$m$$\times$\text{0.1}$m$$\times$\text{0.25}$m$, and the body moves towards the direction of the gripper.
We evaluate the quality of the tracking using the Mean Absolute Error (MAE) of the commanded desired poses and the estimated poses from the kinematics-based state estimator for the manipulator and the torso.
As illustrated in \fig{fig:eef_tracking}, the whole-body controller can enable the robot to accurately track the desired trajectory with tracking MAE of \text{1.06}$mm$ in the end effector position and \text{0.023}$rad$ in the torso orientation.

To further evaluate the tracking performance of the whole-body controller, we define five tasks using the end effector with and without torso movement while standing.
For each task, we define four different intermediate poses for the end effector and the torso. We then generate an 8-second trajectory with total 3200 target 6D poses using cubic Bezier interpolation from each start pose to its end pose.
As illustrated in~\tab{tab:tracking_error}, 
the translational and rotational MAEs of the gripper are only \text{$1.89$}$mm$ and \text{$0.047$}$rad$ respectively even with torso movement, which are much more accurate than learning-based method for foot manipulation~\cite{arm2024pedipulate} with translational error of \text{$57$}$mm$. On the other hand, we notice that the translational errors of the torso during manipulation are much larger than those during standing, which may be due to the uneven deformation of the three stance feet. However, the translational errors of the manipulator are smaller than the torso translational errors, benefiting from the task hierarchy of the whole-body controller.

\subsection{Low-cost Vision-based Teleoperation Platform}

\begin{figure}[t]
	\centering
	\includegraphics[width= \linewidth]{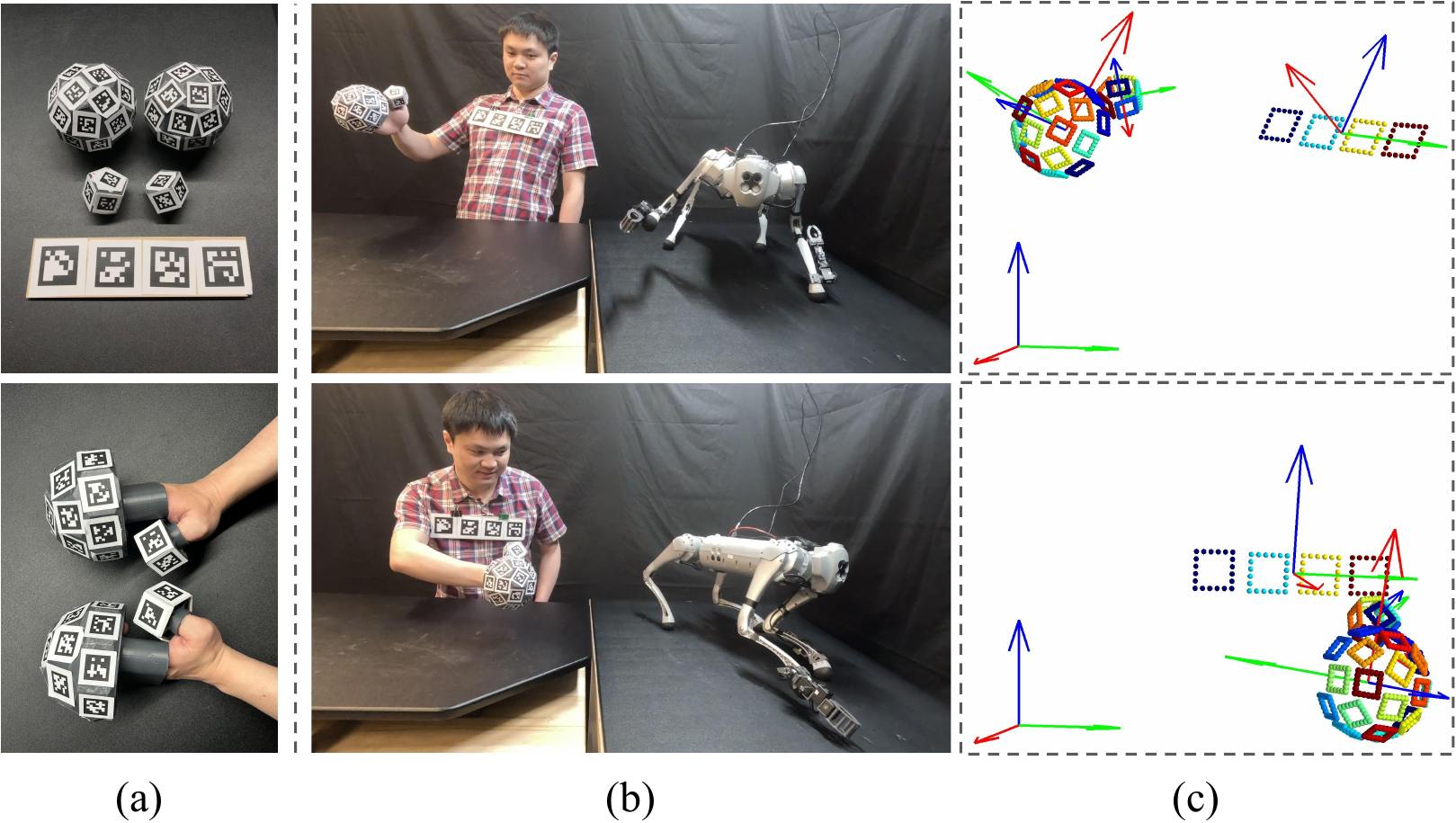}
	\caption{Low-cost teleoperation platform. (a) The rigid shells worn by palms, thumbs and torso created by 3D printing. (b) Simultaneous control of the 6D pose of the torso and the end effector of the quadrupedal robot. (c) Visualization of the estimated corresponding 6D poses of the user's right palm, right thumb and the torso using 3D point clouds of AprilTag edges. }
	\label{fig:teleop}
    \vspace{-9pt}
\end{figure}

In \robot, the robot torso and end effectors need to be commanded simultaneously, which would be difficult to teleoperate using a single joystick, keyboard or sensor glove. 
Therefore, we develop a low-cost vision-based teleoperation platform to teleoperate \robot using motions of human hands and torso.
As illustrated in ~\fig{fig:teleop} (a), the core of our teleoperation platform is a pair of 3D-printed rigid shells worn by palms and thumbs of both hands, as well as the human torso.
The shells are densely covered with AprilTags~\cite{apriltag} whose corners and IDs can be robustly detected.
Using a method similar to \cite{keypose,stereobj:1m}, we set up calibrated high-resolution ZED cameras at multiple views and the 6D pose of the rigid shells can be obtained by minimizing the re-projection errors of AprilTag corners in the multiple-view images.
The 6D pose of the torso and palm shells are directly used to command the 6D pose of the torso and end effector of quadrupedal robot, while the relative angles between the 6D poses of the thumb and palm shells are used to command the opening and closing of the robot gripper.
In this way, the teleoperation users can move their torso and both of their hands and fingers to naturally control all motion of the quadrupedal robot.
With a few cameras and mounts, our teleoperation platform  is especially low-cost in hardware compared to previous works, while allowing the users to simultaneously command torso pose, two end effector poses, and two gripper openings of the quadrupedal robot.

\subsection{Versatile Manipulation in Real-World Tasks}

Using our teleoperation platform, we validate \robot's capabilities listed in~\tab{tab:morphology_comparison} by commanding \robot to complete the following real-world robotic tasks using teleoperation:

\subsubsection{Manipulation with 6DoF End Effector}
To validate that \robot is capable of performing manipulations in the 6D operational space, we evaluate \robot on two challenging tasks: opening a sliding drawer and the swing door of a cabinet. These two tasks demand precise control over the robot end effector pose throughout the interaction.

\begin{figure}[t]
	\centering
	\includegraphics[width= \linewidth]{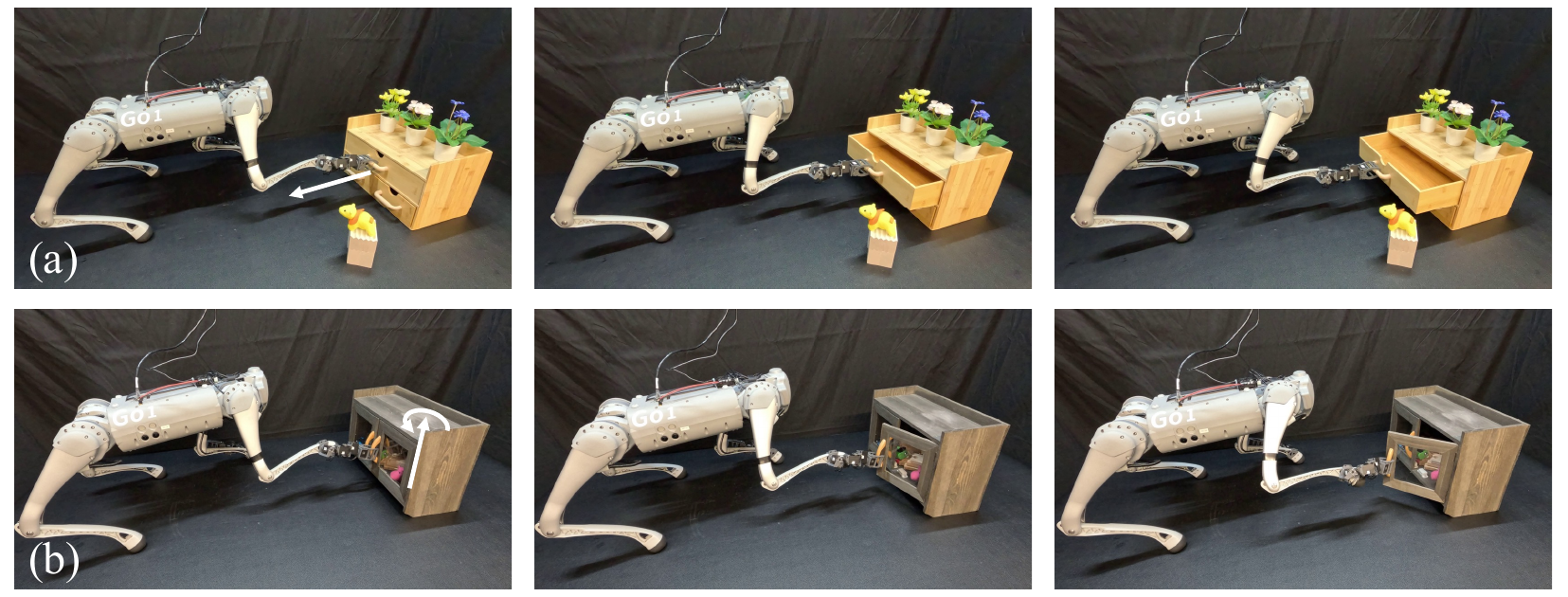}
	\caption{\robot shows its capability of manipulation in 6D operational space in the tasks of opening two typical types of cabinets under vision-based human motion teleoperation. (a) \robot skillfully opens a sliding drawer. (b) \robot adeptly opens a swing door cabinet.}
	\label{fig:cabinet}
\end{figure}

As illustrated in~\fig{fig:cabinet} (a), when opening the sliding drawer, \robot successfully controls the end effector to move linearly along the sliding direction of the drawer while maintaining a stable orientation axis parallel to the movement path. \fig{fig:cabinet}(b) demonstrates \robot's capability in opening a cabinet swing door, where it maneuvers the end effector in an arc trajectory and adjusts the gripper orientation to stay perpendicular to the door plane to avoid collision with the door handle.

\begin{figure}[t]
	\centering
	\includegraphics[width= \linewidth]{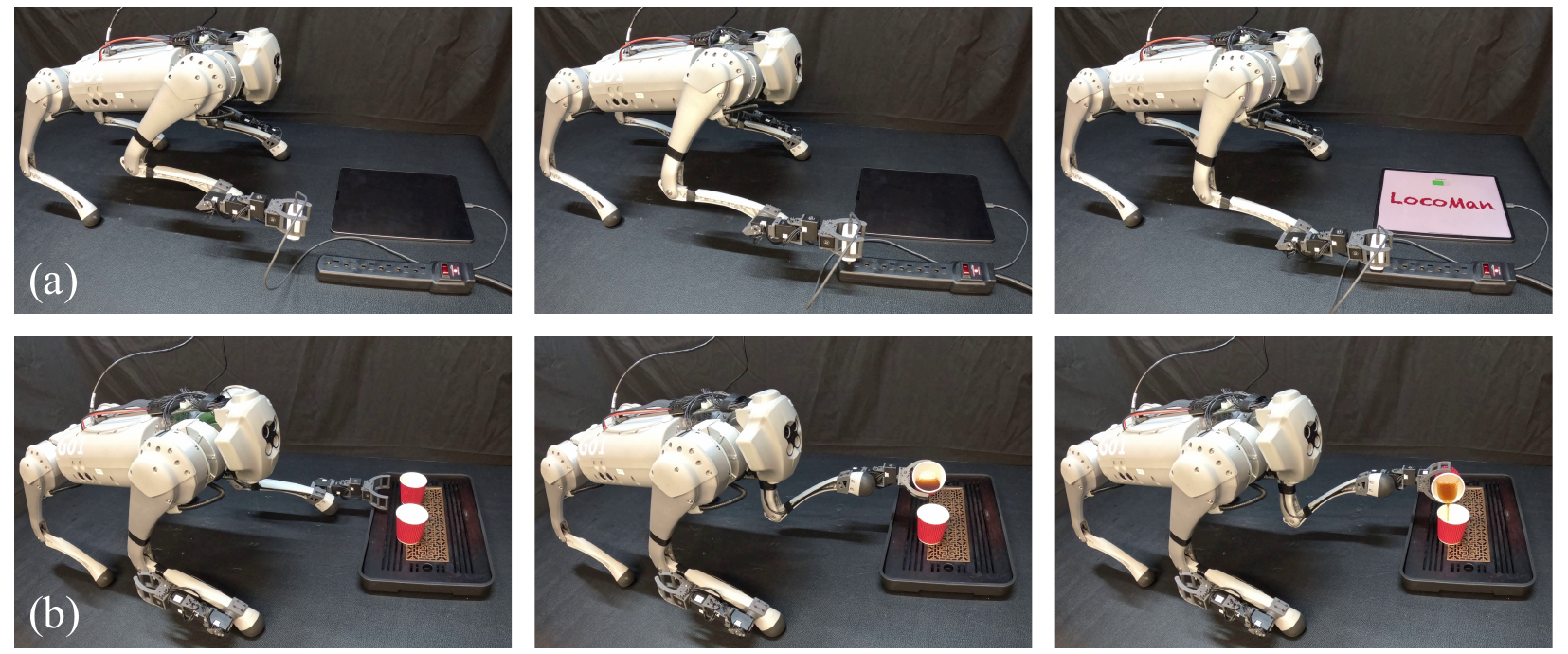}
	\caption{\robot demonstrates its precision and stability in completing two tasks that demand fine control under joystick teleoperation. (a) Insert a charger with meticulous alignment. (b) Pour liquid from a cup to another, carefully modulating the pour angle and rate.}
	\label{fig:usb_coke}
    \vspace{-9pt}
\end{figure}

Furthermore, we teleoperate \robot to perform two additional tasks that require precise motion control: inserting a charger into a power socket shown in~\fig{fig:usb_coke} (a), \robot can accurately align the charger with the socket, demonstrating its ability to precisely adjust its end effector position. In~\fig{fig:usb_coke} (b), \robot pours liquid from one cup to another, which highlights its ability to stably modulate the end effector orientation. 

\subsubsection{Manipulation in Narrow Space}
Fetching objects from an extremely narrow space is a challenge for quadrupedal robots due to the limited space for adjusting both the leg and manipulator joint angles. 

\begin{figure}[t]
	\centering
	\includegraphics[width= \linewidth]{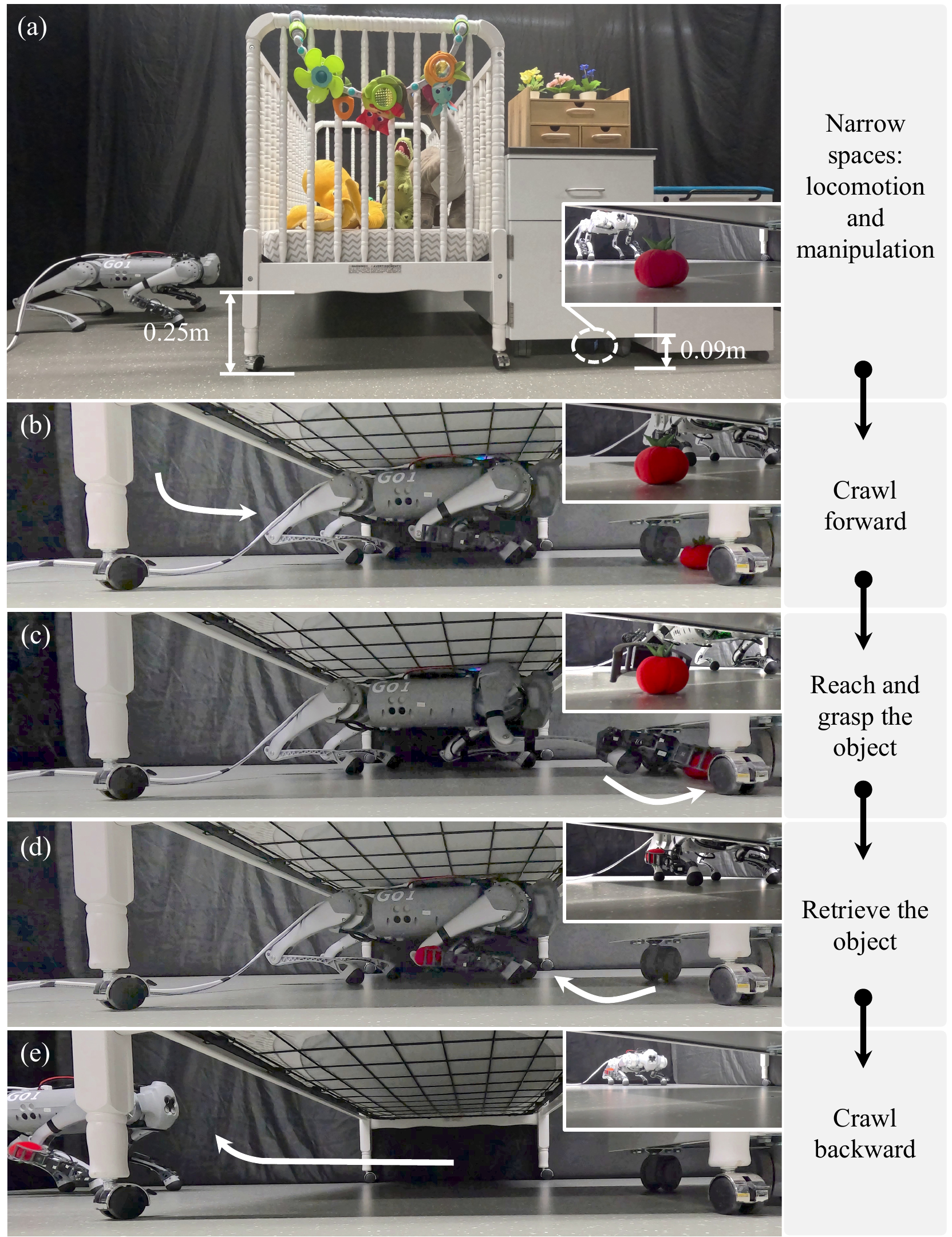}
	\caption{\robot demonstrates its locomotion and manipulation capabilities in constrained environments under joystick teleoperation. (a) The task requires the robot to walk beneath a crib with a clearance of 0.25 m and grasp the object under the cabinet with a clearance of 0.09 m. (b) \robot crawls forward to approach the targeted object. (c) \robot extends its loco-manipulator to reach and grasp the object. (d) \robot retrieves the object. (e) \robot crawls backward and exits the crib.}
	\label{fig:narrow_space}
    \vspace{-12pt}
\end{figure}

As illustrated in~\fig{fig:narrow_space} (a), the task requires \robot to walk below a bed to grasp an object located under a cabinet and then carry it out. This confined environment only offers a \text{$0.25$} $m$ clearance below the bed and an even more constricted \text{$0.09$} $m$ height under the cabinet.
As illustrated in~\fig{fig:narrow_space} (b), \robot can walk stably below the vertically restricted space, avoiding any contact with the bed.
This is only possible without an additional robot arm installed on the top, such as those in \cite{fu2023deep,sleiman2021unified}.
Thanks to the compact design of \manipulator, \robot can deftly maneuver its end effector to extend under the cabinet and adjust its pose to successfully grasp the object.
\robot then retrieves its \manipulator while holding the object in the gripper and exits the narrow space.
The successful completion of this challenging task demonstrates \robot's competency in loco-manipulation within narrow and complex environments. This showcases its potential for deployment in scenarios such as search and rescue missions, which often necessitate both agile locomotion and manipulation in cramped spaces.

\begin{figure}[t]
	\centering
	\includegraphics[width= \linewidth]{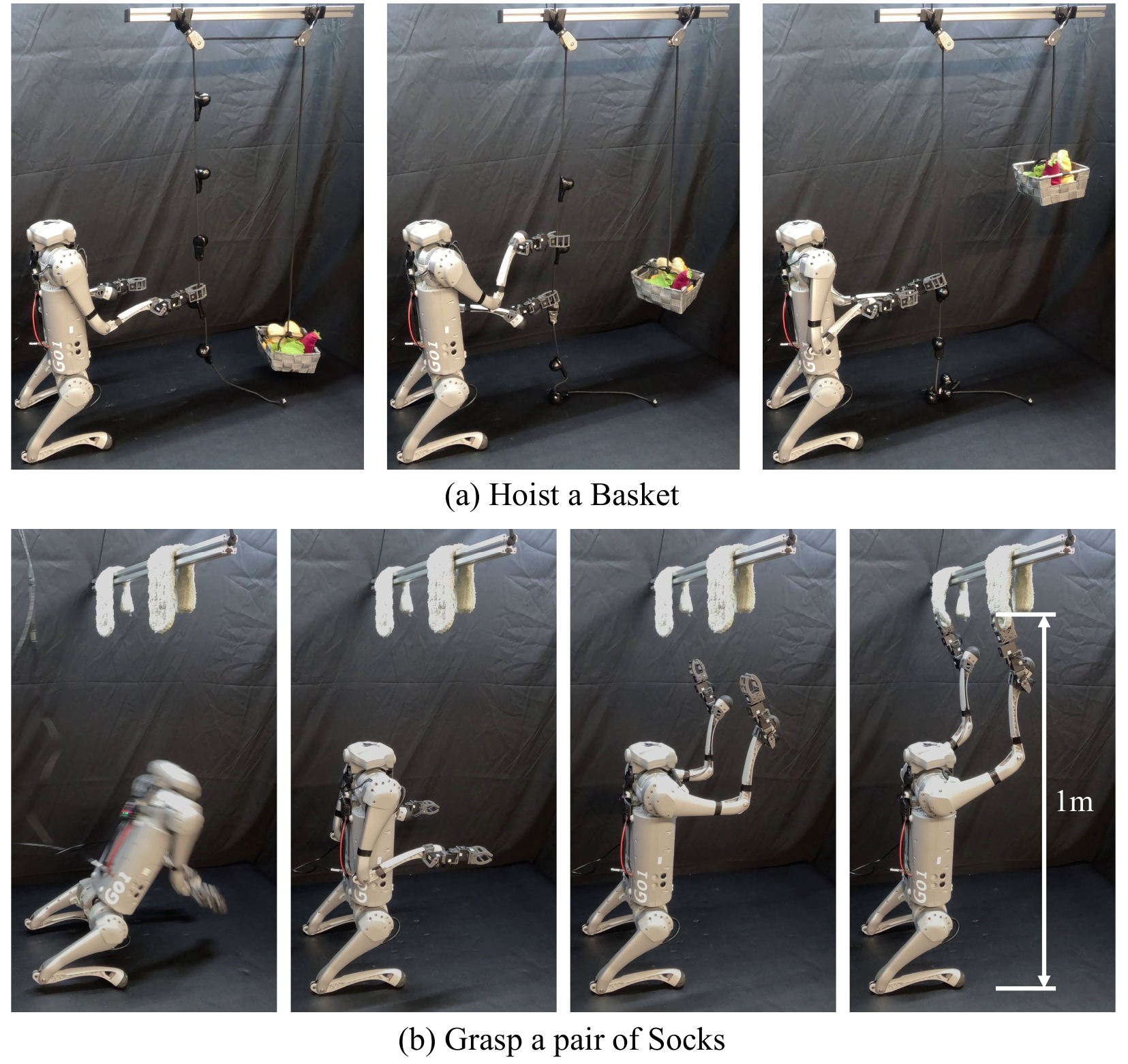}
	\caption{\robot is capable of performing challenging bimanual manipulation tasks with its dual lightweight \manipulators. (a) \robot collaboratively operates its two grippers to hoists a basket under vision-based human motion teleoperation. (b) \robot can reach \text{1}$m$ high to grasp a pair of socks under joystick teleoperation.}
	\label{fig:bimanual}
\end{figure}
\subsubsection{Bimanual Manipulation}
Enabling quadrupedal robots to execute bimanual manipulation significantly broadens the spectrum of tasks they can undertake. 
With dual lightweight \manipulators installed on \robot's two front legs, it becomes practical for \robot to stand upright and engage in complex bimanual tasks.

Hoisting objects with a rope is a task that requires precise bimanual coordination to ensure the rope does not slip back. As illustrated in~\fig{fig:bimanual} (a), while maintaining an upright stance, \robot adeptly adjusts the 6D pose of an end effector to securely grasp the rope and applies a downward pull. Upon reaching the limits of its workspace, the pulling end effector maintains its grip on the rope, steadfastly holding it in place until the alternate end effector takes over to continue pulling the rope. The seamless transition and successful execution of the hoisting operation vividly illustrate \robot's advanced bimanual coordination capabilities, which showcase the practical versatility and enhanced operational scope of quadrupedal robots equipped with our proposed \manipulators.

Furthermore, as~\fig{fig:bimanual} (b) shows, \robot can grasp a pair of socks hanged about \text{$1$} $m$ high from the ground, which is relatively high compared to the height of the robot torso during four legs stance.

\begin{figure}[t]
	\centering
	\includegraphics[width= \linewidth]{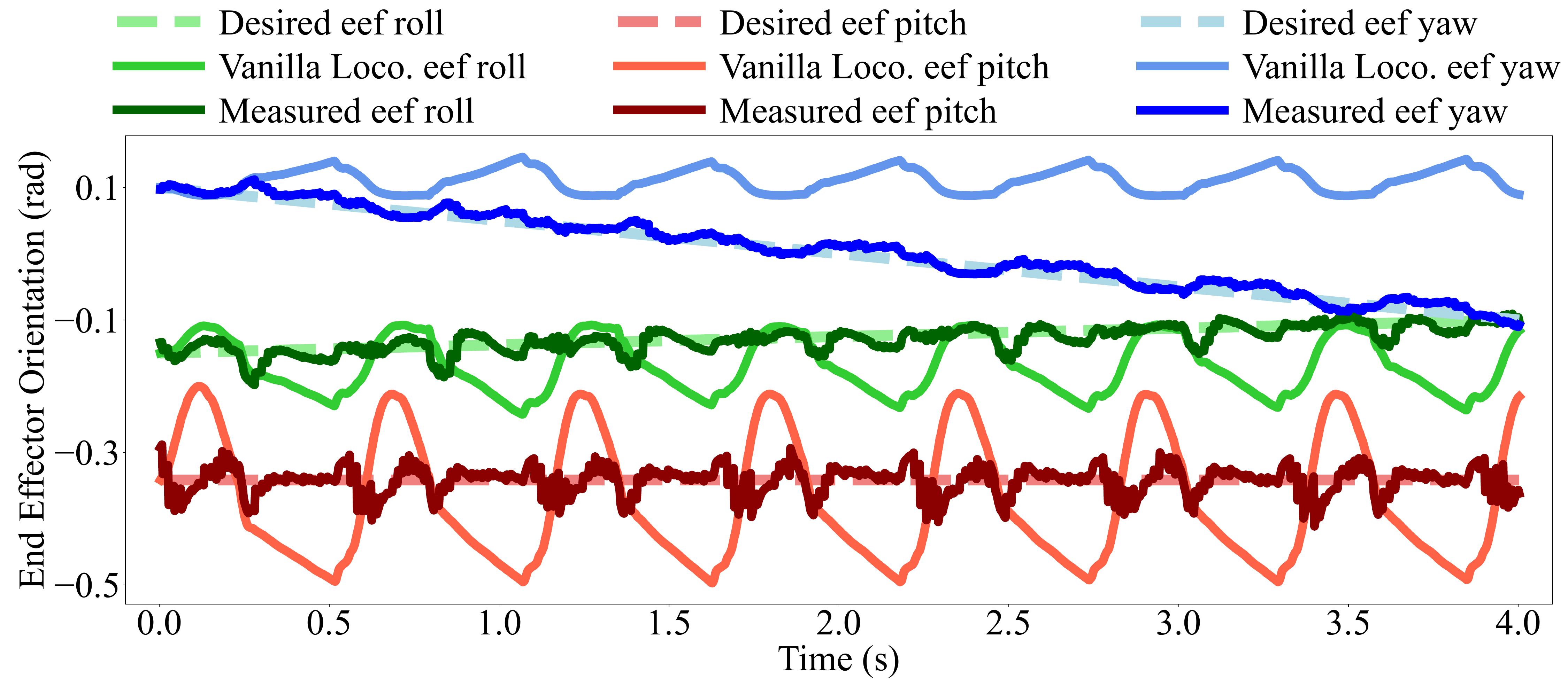}
	\caption{Vanilla Loco. denotes carrying the flashlight using the vanilla locomotion mode. The curves indicate that the loco-manipulation mode of \robot can consistently track the desired orientation of the flashlight in the loco-manipulation mode.}
	\label{fig:loco_manipulation}
        \vspace{-9pt}
\end{figure}
\subsubsection{Loco-Manipulation}

As illustrated in~\fig{fig:teaser} (c), we instruct \robot to manipulate a flashlight aimed at a predetermined orientation trajectory within the world frame, while trotting forward at a velocity of \text{0.15}$m/s$ and shifting left at \text{0.05}$m/s$. This necessitates the coordination of its locomotion and manipulation capabilities. Throughout the execution of this task, we record the orientation of the end effector and also the orientation of its conjunct foot for calculating the vanilla end effector orientation based on their initial relative orientation. As illustrated in~\fig{fig:loco_manipulation}, in the loco-manipulation mode, \robot can track the desired orientation of the flashlight significantly better than relying solely on locomotion.


\section{Conclusion}
\label{sec:conclusion}

	In this paper, we present LocoMan, a novel approach that enhances the manipulation dexterity of quadrupedal robots through the integration of two lightweight loco-manipulators, expanding their operational workspace and enabling precise control over complex 6D manipulation tasks. Our design effectively combines the mobility of quadrupedal robots with the functionality of manipulators, without compromising agility or requiring extensive payload capacity. The developed unified control framework ensures accurate and stable movement across a spectrum of tasks, illustrating LocoMan's versatility in environments ranging from confined spaces to tasks requiring intricate dual-arm coordination.

\section{Acknowledgements}
\label{sec:acknowledgement}
The work is partially supported by Google Deepmind with an unrestricted grant. The authors also acknowledge support from the National Science Foundation under grants CNS-2047454.

\footnotesize{
\bibliographystyle{IEEEtranN}
\bibliography{paper}
}

\newpage

\normalsize
\section*{Appendix}
\label{sec:appendix}
	\subsection{Joint Limits of Loco-Manipulator}
\begin{figure}[ht]
	\centering
	\includegraphics[width= \linewidth]{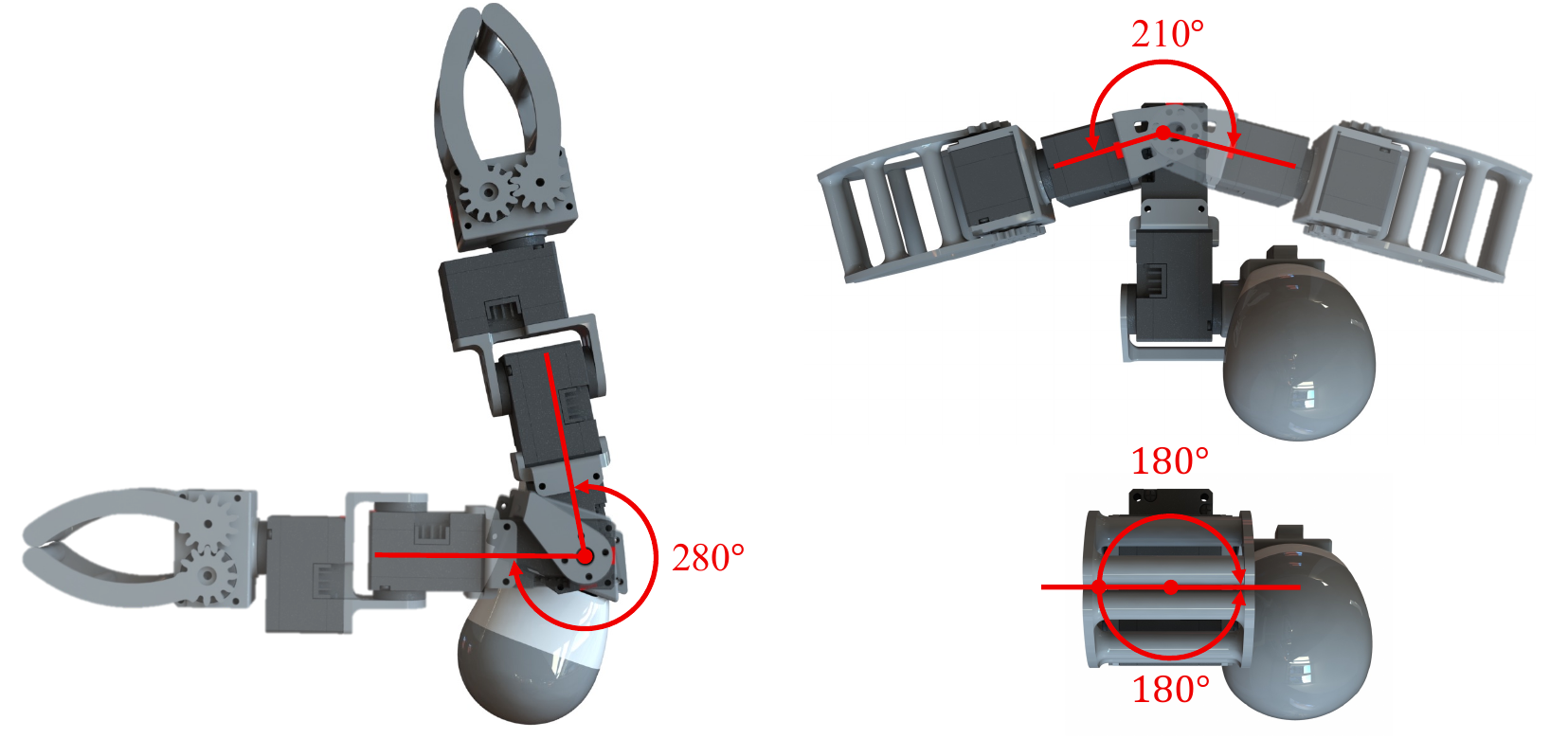}
	\caption{The range of motion for \manipulator's joints. From the first joint to the final one, the corresponding angular movements are $\text{280}$$\degree$, $\text{210}$$\degree$, and $\text{360}$$\degree$ respectively.}
	\label{fig:joint_range}
    \vspace{-8pt}
\end{figure}

The range of motion for \manipulator's joints is illustrated in~\fig{fig:joint_range}, with angles of $\text{280}$$\degree$, $\text{210}$$\degree$, and $\text{360}$$\degree$ for the first through the final joint, respectively.

The motion limitation of the first joint arises from a collision between the first servo and the support structure that is part of the outer calf connector (\fig{fig:design}). This support structure is designed primarily to connect to the opposite horn of the first servo, playing a vital role in stabilizing the manipulator. The design of the support structure aims to minimize its impact on the essential range of motion of the first servo. This critical range is necessary for extending the gripper during transitions from locomotion to manipulation and for positioning the gripper forwards and towards the ground, representing the primary workspace. The limitations in the motion of the second joint are due to self-collision. For the third joint, it does not face stringent motion restrictions, except for those imposed by the cable's length.

\subsection{Whole-Body Impulse Control}
\textbf{Notation} from~\ref{sec:whole_body}: robot state $\vec{q}=(\vec{x}^{\text{torso}}, \vec{q}_{j})\in \mathbb{R}^{24}$, where $\vec{x}^{\text{torso}}\in\mathbb{R}^{6}$ denotes the position and orientation of the floating torso, and $\vec{q}_{j} \in\mathbb{R}^{18}$ denotes the joint angles of \robot. The bar notation $(\overline{\cdot})$ denotes all desired states.

\subsubsection{\textbf{Hierarchical Objective Tracking}}
As depicted in~\tab{tab:operation_modes}, there are $n$ tracking objectives in different operation modes, comprising $n$ tracking tasks to be executed. The null space projection technique is utilized to track each objective in hierarchical order (from $1$-st to $n$-th).

In the $i$-th ($i=1, ..., n$) task, the tracking items consist of the desired position $\overline{\vec{x}_{i}}$, velocity $\overline{\dot{\vec{x}}_{i}}$, and acceleration $\overline{\ddot{\vec{x}}_{i}}$ of the $i$-th tracking objective, which are tracked by computing the desired position $\overline{\vec{q}}_{i}$, velocity $\overline{\dot{\vec{q}}}_{i}$, and acceleration $\overline{\ddot{\vec{q}}}_{i}$ for both the floating torso and the joints. The desired states ($\overline{\vec{q}}_{n}$, $\overline{\dot{\vec{q}}}_{n}$, and $\overline{\ddot{\vec{q}}}_{n}$) for the last tracking objective would be the final desired states ($\overline{\vec{q}}$, $\overline{\dot{\vec{q}}}$, and $\overline{\ddot{\vec{q}}}$) of \robot.

\begin{equation}
\begin{aligned}
\overline{\vec{q}_{i}} & = \overline{\vec{q}_{i-1}} + {\boldsymbol{J}_{i\mid pre}}^{\dagger}\left(\overline{\vec{x}_{i}}-\vec{x}_{i} - \boldsymbol{J}_{i}\left(\overline{\vec{q}_{i-1}}-\vec{q}\right)\right)\\
\overline{\dot{\vec{q}}_{i}} & = \overline{\dot{\vec{q}}_{i-1}} + {\boldsymbol{J}_{i \mid pre}}^{\dagger}\left(\overline{\dot{\vec{x}}_{i}} - \boldsymbol{J}_{i}\overline{\dot{\vec{q}}_{i-1}}\right) \\
\overline{\ddot{\vec{q}}_{i}} & = {\boldsymbol{J}_{i\mid pre}^{\mathrm{dyn}}}^{\ddagger}\left(\ddot{\vec{x}}_{i}^{\mathrm{cmd}} - \dot{\boldsymbol{J}}_{i} \dot{\vec{q}}-\boldsymbol{J}_{i} \overline{\ddot{\vec{q}}_{i-1}}\right)
\end{aligned}
\label{task_hierarchy}
\end{equation}
The acceleration command $\ddot{\vec{x}}_{i}^{\mathrm{cmd}}$ is computed by:
\begin{equation}
\ddot{\vec{x}}_{i}^{\mathrm{cmd}}=\overline{\ddot{\vec{x}}_{i}}+\boldsymbol{K}_{p}^{acc}\left(\overline{\vec{x}_{i}}-\vec{x}_{i}\right)+\boldsymbol{K}_{d}^{acc}\left(\overline{\dot{\vec{x}}_{i}}-\dot{\vec{x}}_{i}\right)
\label{acceleration_command}
\end{equation}
where $\boldsymbol{K}_{p}^{acc}$ and $\boldsymbol{K}_{d}^{acc}$ denote position and velocity feedback gains.

To compute the desired states of \robot from the tracking objective, we need to use the pseudo-inverse of the Jacobian. In~\eq{task_hierarchy}, we track the desired position and velocity kinematically, and the desired acceleration dynamically. Therefore, two types of pseudo-inverses are utilized. One is the SVD-based pseudo-inverse denoted by $\{\cdot\}^{\dagger}$, whcih is defined as:
\begin{equation}
\begin{aligned}
\boldsymbol{J}= \boldsymbol{U}\boldsymbol{\Sigma}\boldsymbol{V}^T \\
\boldsymbol{J}^{\dagger} = \boldsymbol{V}\boldsymbol{\Sigma}^{\dagger}\boldsymbol{U}^T
\end{aligned}
\label{svd_inverse}
\end{equation}
where $\boldsymbol{J}$ is the Jacobian and $\boldsymbol{U}$ and $\boldsymbol{V}$ are the orthogonal matrices. Besides, $\boldsymbol{\Sigma}$ is the diagonal matrix, and $\boldsymbol{\Sigma}^{\dagger}$ is formed by taking the reciprocal of each non-zero element on the diagonal of $\boldsymbol{\Sigma}$, leaving the zeros in place, and then transposing the matrix.

The other is the dynamically consistent pseudo-inverse denoted by $\{\cdot\}^{\ddagger}$ with the definition of:
\begin{equation}
\begin{aligned}
\boldsymbol{J}^{\ddagger}=\boldsymbol{A}^{-1} \boldsymbol{J}^{\top}\left(\boldsymbol{J} \boldsymbol{A}^{-1} \boldsymbol{J}^{\top}\right)^{-1} 
\end{aligned}
\label{dynamical_inverse}
\end{equation}
where $\boldsymbol{A}$ is the mass matrix.

Furthermore, $\boldsymbol{J}_{i}$ is the Jacobian of the $i$-th tracking objective, while $\boldsymbol{J}_{i\mid pre}$ is the projection of it into the null space of all previous tasks. And the symbol $\{\cdot\}^{\mathrm{dyn}}$ denotes the dynamically consistent constraints. $\boldsymbol{J}_{i\mid pre}$ and $\boldsymbol{J}_{i\mid pre}^{\mathrm{dyn}}$ are computed with the following steps:

\begin{equation}
\begin{aligned}
& \boldsymbol{N}_{i-1}=\boldsymbol{N}_{0} \boldsymbol{N}_{1 \mid 0} \cdots \boldsymbol{N}_{i-1 \mid i-2} \\
& \boldsymbol{J}_{i\mid pre}=\boldsymbol{J}_{i} \boldsymbol{N}_{i-1} \\
& \boldsymbol{J}_{i \mid i-1}= \boldsymbol{J}_{i} \left( \boldsymbol{I}-{\boldsymbol{J}_{i-1}}^{\dagger} \boldsymbol{J}_{i-1} \right) \\    
& \boldsymbol{N}_{i \mid i-1}=\boldsymbol{I}-{\boldsymbol{J}_{i \mid i-1}}^{\dagger} \boldsymbol{J}_{i \mid i-1} \\
& \boldsymbol{N}_{i-1}^{\mathrm{dyn}}=\boldsymbol{N}_{0}^{\mathrm{dyn}} \boldsymbol{N}_{1 \mid 0}^{\mathrm{dyn}} \cdots \boldsymbol{N}_{i-1 \mid i-2}^{\mathrm{dyn}} \\
& \boldsymbol{J}_{i\mid pre}^{\mathrm{dyn}}=\boldsymbol{J}_{i} \boldsymbol{N}_{i-1}^{\mathrm{dyn}} \\
& \boldsymbol{J}_{i \mid i-1}^{\mathrm{dyn}}= \boldsymbol{J}_{i} \left( \boldsymbol{I}-{\boldsymbol{J}_{i-1}}^{\ddagger} \boldsymbol{J}_{i-1} \right) \\
& \boldsymbol{N}_{i \mid i-1}^{\mathrm{dyn}}=\boldsymbol{I}-{\boldsymbol{J}_{i \mid i-1}^{\mathrm{dyn}}}^{\ddagger} \boldsymbol{J}_{i \mid i-1}^{\mathrm{dyn}} \\
\end{aligned}
\label{projection}
\end{equation}

where $\boldsymbol{N}_{i-1}$, $\boldsymbol{J}_{i \mid i-1}$, and $\boldsymbol{N}_{i \mid i-1}$ represent the null space of the previous ($i-1$) tasks, the projection of the Jacobian $\boldsymbol{J}_{i}$ into the null space of the ($i-1$)-th task, and the independent null space introduced by the $i$-th task. 

Here, $i$ is within the range of $1, ..., n$. And the initial items are defined as:
\begin{equation}
\begin{aligned}
& \boldsymbol{N}_{0}=\boldsymbol{I}-{\boldsymbol{J}_{c}}^{\dagger } \boldsymbol{J}_{c} \\
& \boldsymbol{N}_{0}^{\mathrm{dyn}}=\boldsymbol{I}-{\boldsymbol{J}_{c}}^{\ddagger} \boldsymbol{J}_{c} \\
& \boldsymbol{J}_{0} = \boldsymbol{J}_{c} \\
& \overline{\vec{q}_{0}} = \vec{q}\\
& \overline{\dot{\vec{q}}_{0}} = \vec{0}\\
& \overline{\ddot{\vec{q}}_{0}}= {\boldsymbol{J}_{c}}^{\ddagger} \left(-\boldsymbol{J}_{c} \dot{\vec{q}}\right)
\end{aligned}
\end{equation}
where $\boldsymbol{J}_{c}$ is the Jacobian of the feet in contact.

\subsubsection{\textbf{Quadratic Programming}}
With the desired acceleration $\overline{\ddot{\vec{q}}}$ computed in the previous step, we solve a Quadratic Program (QP) to optimize for ground reaction forces.
The QP problem is formulated as:
\begin{equation}
\begin{aligned}
\min _{\mathbf{f}_{r}, \boldsymbol{\delta}_{t}} \mathbf{f}_{r}^{\top} \boldsymbol{Q}_{1} \mathbf{f}_{r}+\boldsymbol{\delta}_{t}^{\top} \boldsymbol{Q}_{2} \boldsymbol{\delta}_{t}
\end{aligned}
\end{equation}
\begin{align}
\text{s.t. } & \boldsymbol{S}_{f}(\boldsymbol{A} \ddot{\mathbf{q}}^{\mathrm{cmd}} +\mathbf{b}+\mathbf{g})=\boldsymbol{S}_{f} \boldsymbol{J}_{c}^{\top} \mathbf{f}_{r} \tag{floating torso dyn.}\\
& \ddot{\mathbf{q}}^{\mathrm{cmd}}=\overline{\ddot{\vec{q}}}+\begin{bmatrix}
\boldsymbol{\delta}_{t} \\
\mathbf{0}_{j}
\end{bmatrix} \tag{floating torso acceleration}\\
& \boldsymbol{W} \mathbf{f}_{r} \geq \mathbf{0} \tag{contact force constraints}
\end{align}
where the optimization objective is to minimize the reaction force $\mathbf{f}_{r}$ and the relaxation variable $\boldsymbol{\delta}_{t}$ for the floating torso acceleration with the weights $\boldsymbol{Q}_{1}$ and $\boldsymbol{Q}_{2}$.  $\boldsymbol{S}_{f}$, $\boldsymbol{A}$, $\mathbf{b}$, $\mathbf{g}$, and $\boldsymbol{W}$ are the floating torso selection matrix, the mass matrix, the Coriolis force, the gravitation force, and the force constraint matrix for foot contact, respectively.

With the optimized reaction force $\mathbf{f}_{r}$ and relaxation variable $\boldsymbol{\delta}_{t}$ for the floating torso acceleration, we compute the joint torque command with the multi-body dynamics formulation:
\begin{equation}
\left[\begin{array}{c}
\overline{\vec{\tau}_{t}}\\
\overline{\vec{\tau}_{j}}
\end{array}\right]=A \ddot{\mathbf{q}}^{\mathrm{cmd}}+\mathbf{b}+\mathbf{g}-\boldsymbol{J}_{c}^{\top} \mathbf{f}_{r}
\end{equation}
where $\boldsymbol{\tau}_{j}$ is the torque command for \robot's joints.

\subsubsection{\textbf{Joint-Level PD Control}}
We compute the desired position $\overline{\vec{q}_{j}}$, velocity $\overline{\dot{\vec{q}}_{j}}$, and torque $\overline{\vec{\tau}_{j}}$ for \robot's joints in the previous steps, which runs in \text{400}Hz. The final torque command $\vec{\tau}_{j}$ applied to each joint is computed by a joint-level PD controller that runs in \text{2000}Hz:
\begin{equation}
\vec{\tau}_{j} = \overline{\vec{\tau}_{j}} + \boldsymbol{K}_{p}^{tau}\left(\overline{\vec{q}_{j}}-\vec{q}_{i}\right)+\boldsymbol{K}_{d}^{tau}\left(\overline{\dot{\vec{q}}_{j}}-\dot{\vec{q}}_{i}\right)
\end{equation}
The high frequency joint-level PD feedback enables the robot to track desired poses smoothly and accurately.

\subsubsection{\textbf{PD Gains for Real Robot}}
In the experiments described in~\ref{sec:experiments}, we tune the PD gains of the tracking objectives for the whole-body impulse controller (~\tab{tab:wbc_pd}) and the PD gains of the hip, thigh, and calf joints for the joint-level PD controller (~\tab{tab:joint_pd}). We use the SDK from Dynamixel\textsuperscript{\textregistered} and set the position gain values to ${\left[8, 6, 6\right]}^{T} \times 128$ and the velocity gain values to ${\left[65, 40, 40\right]}^{T} \times 16$ for the three joints of \manipulator in all operation modes, where $128$ and $16$ are the default scales in the SDK.

\begin{table}[ht]
\centering
\caption{PD gains of the tracking objectives for computing the acceleration errors in whole-body impulse control.}
\begin{tabular}{cccc}\toprule
Tracking Objective &  $\boldsymbol{K}_{p}^{acc}$ &  $\boldsymbol{K}_{d}^{acc}$ \\
\midrule
Torso Position (Locomotion) & ${\left[100, 100, 100\right]}^{T}$ & ${\left[10, 10, 10\right]}^{T}$ \\
Torso Orientation (Locomotion) & ${\left[100, 100, 100\right]}^{T}$ & ${\left[10, 10, 10\right]}^{T}$ \\
Torso Position (Manipulation) & ${\left[100, 100, 100\right]}^{T}$ & ${\left[1, 1, 1\right]}^{T}$ \\
Torso Orientation (Manipulation) & ${\left[100, 100, 100\right]}^{T}$ & ${\left[1, 1, 1\right]}^{T}$ \\
Foot Position & ${\left[100, 100, 100\right]}^{T}$ & ${\left[10, 10, 10\right]}^{T}$ \\
Foot Orientation & ${\left[100, 100, 100\right]}^{T}$ & ${\left[10, 10, 10\right]}^{T}$ \\
Gripper Position & ${\left[100, 100, 100\right]}^{T}$ & ${\left[10, 10, 10\right]}^{T}$ \\
Gripper Orientation & ${\left[100, 100, 100\right]}^{T}$ & ${\left[10, 10, 10\right]}^{T}$ \\
\bottomrule
\end{tabular}
\label{tab:wbc_pd}
\end{table}

\begin{table}[ht]
\setlength{\tabcolsep}{1pt}
\centering
\caption{PD gains of the hip, thigh, and calf joints for computing the applied torque in joint-level PD control.}
\begin{tabular}{cccc}\toprule
Leg Mode &  $\boldsymbol{K}_{p}^{tau}$ &  $\boldsymbol{K}_{d}^{tau}$ \\
\midrule
Stance Legs in Locomotion & ${\left[30, 30, 30\right]}^{T}$ & ${\left[1, 1, 1\right]}^{T}$ \\
Swing Legs in Locomotion & ${\left[30, 30, 30\right]}^{T}$ & ${\left[1, 1, 1\right]}^{T}$ \\
Stance Legs in Loco-Manipulation & ${\left[30, 30, 30\right]}^{T}$ & ${\left[1, 1, 1\right]}^{T}$ \\
Swing Legs in Loco-Manipulation & ${\left[20, 20, 20\right]}^{T}$ & ${\left[0.8, 0.8, 0.8\right]}^{T}$ \\
Stance Legs in Single-Arm Manipulation & ${\left[60, 60, 60\right]}^{T}$ & ${\left[2, 2, 2\right]}^{T}$ \\
Swing Legs in Single-Arm Manipulation & ${\left[60, 60, 60\right]}^{T}$ & ${\left[2, 2, 2\right]}^{T}$ \\
Stance Legs in Bimanual Manipulation & ${\left[100, 100, 100\right]}^{T}$ & ${\left[2.5, 2.5, 2.5\right]}^{T}$ \\
Swing Legs in Bimanual Manipulation & ${\left[30, 30, 30\right]}^{T}$ & ${\left[1, 1, 1\right]}^{T}$ \\
\bottomrule
\end{tabular}
\label{tab:joint_pd}
\end{table}

\subsection{State Estimation}
As illustrated in~\tab{tab:operation_modes}, there are two state estimators in our system, including the kinematics-based estimator used in the manipulation modes and the Kalman filter-based estimator used in the locomotion mode and the loco-manipulation mode.

\begin{figure}[ht]
	\centering
	\includegraphics[width= \linewidth]{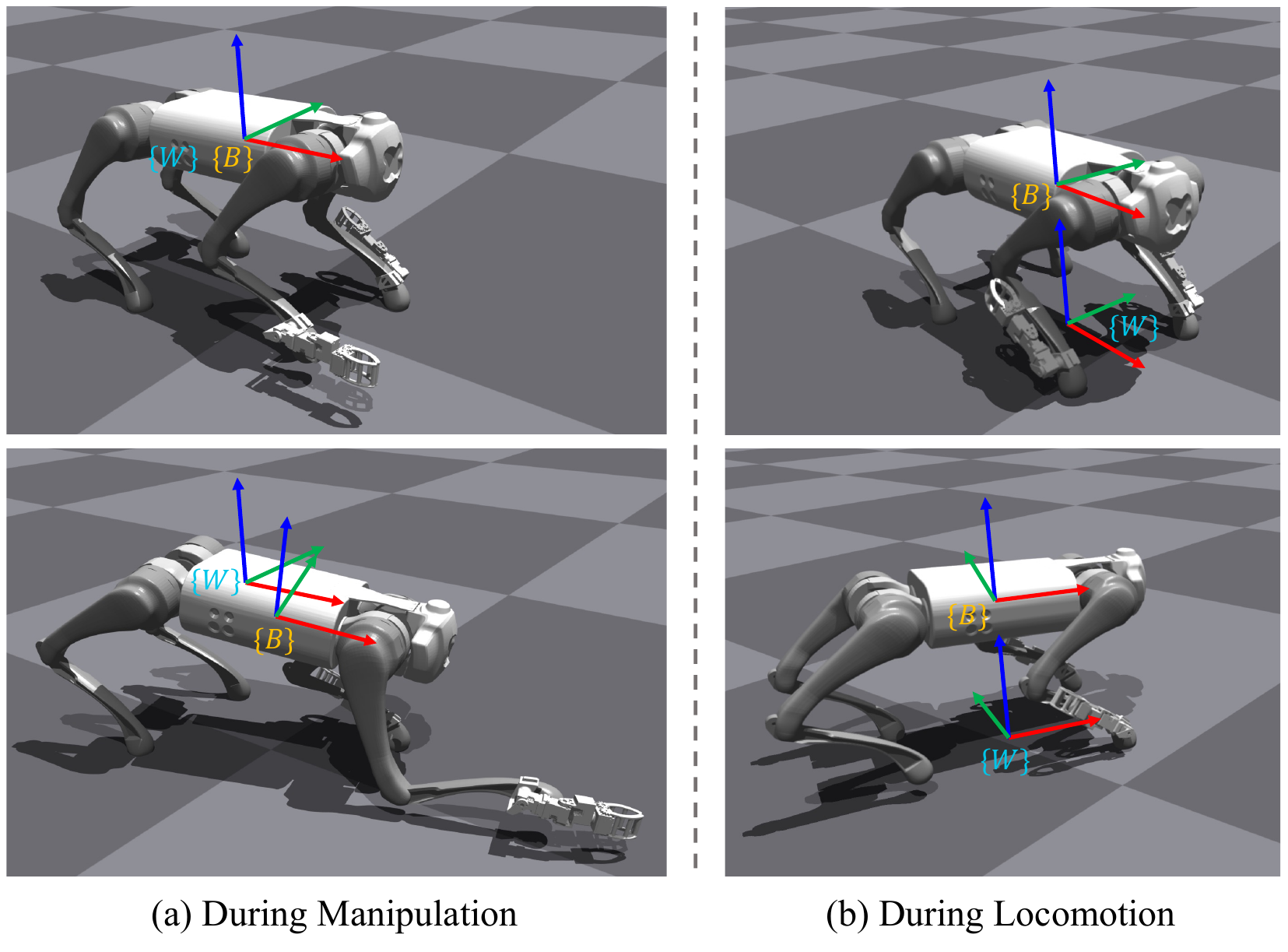}
	\caption{The definition of the world frame in different state estimators. (a) The kinematics-based estimator set the initial base frame of \robot as the world frame. (b) The Kalman filter-based estimator always has the zero-yaw world frame as the world frame.}
	\label{fig:world_frame}
    \vspace{-8pt}
\end{figure}

As depicted in~\fig{fig:world_frame} (a), at the moment when \robot finishes the transition process from other operation modes to manipulation modes, the kinematics-based estimator sets the base frame of \robot as the world frame. Furthermore, we also record the positions of the contact foot as $\vec{P}_{init}^{W}$, which can be computed using forward kinematics and have the same values in the world frame or the base frame at the reset moment. During manipulation, the world frame remains fixed, and we compute the current positions of the contact foot in the base frame as $\vec{P}_{cur}^{B}$. Assuming that there is no slip occurring between the contact foot and the ground, we employ the Procrustes analysis method to solve the transform matrix from the world frame to the base frame $\vec{T}_{W}^{B}$.

As for locomotion, we implement the Kalman filter-based estimator which fuses the IMU information. As~\fig{fig:world_frame} (b) shows, the world frame is always updated to be the zero-yaw world frame.

\end{document}